\documentclass[runningheads]{llncs}
\usepackage[T1]{fontenc}
\usepackage{graphicx}
\usepackage{graphicx}
\usepackage[utf8]{inputenc}
\usepackage{caption}
\usepackage{subcaption}
\usepackage{booktabs}
\usepackage{multirow, makecell}
\usepackage{breakcites} 
\usepackage{pifont}

\usepackage{hyperref}
\hypersetup{
    colorlinks=true,
    linkcolor=blue,
    filecolor=magenta,      
    urlcolor=magenta,
    pdftitle={Masked Self-Supervised Pre-Training for Text Recognition Transformers on Large-Scale Datasets},
    pdfpagemode=FullScreen,
}

\usepackage[english]{babel}
\usepackage{xcolor,colortbl}
\usepackage{blindtext}

\begin{document}
\title{Masked Self-Supervised Pre-Training for Text Recognition Transformers on Large-Scale Datasets}
\titlerunning{Masked Self-Supervised Pre-Training for Text Recognition}
%
\author{
Martin Kišš\inst{1} \orcidID{0000-0001-6853-0508} \and
Michal Hradiš\inst{1} \orcidID{0000-0002-6364-129X}
}
\authorrunning{M. Kišš et al.}
%
\institute{
Faculty of Information Technology, Brno University of Technology, \\
Brno, Czech Republic\\
\email{\{ikiss,hradis\}@fit.vutbr.cz}
}
\maketitle              
\begin{abstract}
Self-supervised learning has emerged as a powerful approach for leveraging large-scale unlabeled data to improve model performance in various domains.
In this paper, we explore masked self-supervised pre-training for text recognition transformers.
Specifically, we propose two modifications to the pre-training phase: progressively increasing the masking probability, and modifying the loss function to incorporate both masked and non-masked patches.
We conduct extensive experiments using a dataset of 50M unlabeled text lines for pre-training and four differently sized annotated datasets for fine-tuning.
Furthermore, we compare our pre-trained models against those trained with transfer learning, demonstrating the effectiveness of the self-supervised pre-training.
In particular, pre-training consistently improves the character error rate of models, in some cases up to 30 \% relatively.
It is also on par with transfer learning but without relying on extra annotated text lines.

\end{abstract}
\section{Introduction}
Large models have become fundamental in machine learning, with notable examples including large language models~\cite{brown_language_2020,openai_gpt-4_2024,grattafiori_llama_2024}, the ASR model Whisper~\cite{radford_robust_2022}, and generative models like Stable Diffusion~\cite{rombach_high-resolution_2022,esser_scaling_2024} and DALLE-3~\cite{betker_improving_nodate}.
Such large models require huge amounts of training data to generalize well.
In text recognition, and especially in handwritten text recognition, acquiring large datasets containing millions of text lines is costly and time-consuming.
On the other hand, there are machine learning paradigms that utilize unlabeled data to overcome this issue, namely semi-supervised and self-supervised learning.

Approaches to semi-supervised learning most commonly use pseudo-labeling, also called self-training,~\cite{lee_pseudo-label_2013,kiss_at-st_2021,kiss_softctcsemi-supervised_2023,xie_self-training_2020} and consistency regularization~\cite{aberdam_multimodal_2022,berthelot_mixmatch_2019,kurakin_remixmatch_2020}.
In text recognition, pseudo-labeling uses unlabeled text lines annotated with machine transcriptions to improve an existing system, while consistency regularization is based on the idea that a text line should be transcribed the same under different perturbations.
The self-supervised learning is typically used to pre-train a model on unlabeled data and it is then fine-tuned for a downstream task, like text recognition, speech recognition, or image classification.
Self-supervised approaches are mainly based on masked label prediction~\cite{hsu_hubert_2021,bao_beit_2022,peng_beit_2022,han_nest-rq_2024,chiu_self-supervised_2022,assran_self-supervised_2023} and joint-embedding learning~\cite{bardes_vicreg_2022,bardes_vicregl_2022,chen_simple_2020}.
In the masked label prediction, random parts of the input data are masked and the model is trained to predict correct labels of the masked parts.
The joint-embedding learning uses two views of the input data that are processed by a model which is trained to produce similar outputs for the corresponding outputs while distancing the unrelated ones.

\begin{figure}[t]
    \centering
    \includegraphics[width=\linewidth]{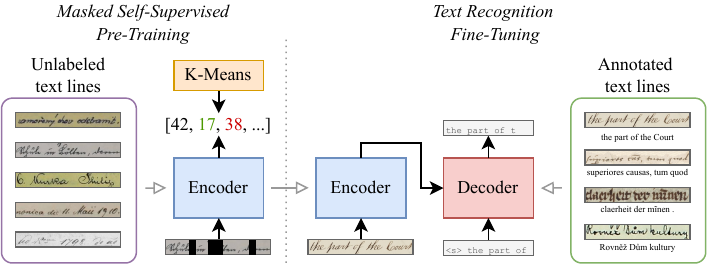}
    \caption{
    Overview of our self-supervised pre-training.
    First, we pre-train encoder using fitted K-Means and masked label prediction.
    Then, we use the encoder in encoder-decoder model and we fine-tune it on an annotated dataset.
    }
    \label{fig:intro}
\end{figure}

In this paper, we investigate the self-supervised pre-training, specifically the masked label prediction utilizing K-Means clustering~\cite{hsu_hubert_2021,kiss_self-supervised_2024} (see Figure~\ref{fig:intro}).
We propose two modifications to the pre-training phase which we experimentally evaluate.
We also experiment with two fine-tuning strategies and six differently sized models.
Within the experiments, we use dataset of 50M unlabeled text lines for the pre-training and four datasets of different sizes for fine-tuning -- three standard, publicly available datasets (the Bentham collection~\cite{sanchez_icfhr2014_2014}, the Bullinger dataset~\cite{scius-bertrand_bullinger_2023}, and the CATMuS Medieval dataset~\cite{clerice_catmus_2024}) and an internal handwritten dataset called PERO-HWR.
The sizes of the fine-tuning datasets range from 9k to over 1M annotated text lines.
We compare our pre-trained models also with models trained using transfer learning, which is usually considered a very strong baseline~\cite{kiss_self-supervised_2024}.

The contributions of the paper are as follows:
\begin{enumerate}
    \item We investigate two proposed modifications of the masked label prediction-based self-supervised pre-training on a large-scale unlabeled dataset.
    \item We evaluate two fine-tuning strategies on four differently sized annotated datasets.
    \item We experiment with six differently sized pre-trained models, we compare them to models trained using transfer learning and we also compare them with the state-of-the-art results.
\end{enumerate}

\section{Related work}
In recent years, self-supervised learning has become a prominent approach in machine learning, enabling large Transformer-based models to be trained without relying on extensive labeled datasets. 
This paradigm encourages models to learn meaningful representations by generating supervision signals from the input data.
Self-supervised learning has been successfully applied to various domains, including language modeling~\cite{devlin_bert_2018,baevski_efficient_2023}, image modeling~\cite{bao_beit_2022,caron_emerging_2021,chen_simple_2020,zhou_ibot_2022,bardes_vicreg_2022}, and automatic speech recognition~\cite{hsu_hubert_2021,chiu_self-supervised_2022,liu_uniwav_2024,chen_wavlm_2022}.

A common technique in self-supervised learning is masked label prediction~\cite{kiss_at-st_2021,hsu_hubert_2021,bao_beit_2022}, where parts of the input or intermediate representation are masked, and the model is trained to infer the missing information based on its context. 
In this setup, the labels are not the actual annotations from the downstream task but are derived from the input data itself, ensuring that similar patterns receive consistent labels. 
By leveraging this method, the model learns dependencies between the masked elements and their surrounding context, ultimately improving its ability to represent data.
The masking-based modeling approach was first introduced in the pre-training of Transformer-based language model BERT~\cite{devlin_bert_2018}.

The second commonly used technique is joint-embedding learning, where the model processes two different views of the input data and optimizes their outputs to be similar for corresponding pairs while pushing apart all other outputs.
This approach relies on specialized loss functions, such as VICReg~\cite{bardes_vicreg_2022,bardes_vicregl_2022} or NT-Xent~\cite{chen_simple_2020}, to achieve effective representation learning.

\subsection{Speech recognition}
One of the earliest self-supervised ASR approaches is wav2vec 2.0~\cite{baevski_wav2vec_2020} by Baevski et al., which processes raw audio through a convolutional encoder to produce a latent representation. 
Randomly masked parts are refined by a Transformer encoder, while product quantization is applied. 
The model is trained via contrastive loss, distinguishing masked representations from quantized positives and negatives.
HuBERT~\cite{hsu_hubert_2021} by Hsu et al. follows a similar pipeline but generates training targets by clustering Mel-frequency cepstral coefficients (MFCCs) with K-Means. WavLM~\cite{chen_wavlm_2022} builds on HuBERT, introducing noise augmentation for speaker recognition, separation, and diarization.
Another masked training approach, BEST-RQ~\cite{chiu_self-supervised_2022}, generates labels using a frozen randomly initialized projection and codebook.
Similarly, in NEST-RQ~\cite{han_nest-rq_2024} the labels are generated using a frozen randomly initialized projection and codebook, but the model is trained using next token prediction.
In UniWav~\cite{liu_uniwav_2024}, a student encoder is trained with masked audio modeling and pseudo-labels obtained using a teacher encoder model.
The student is trained using standard backpropagation and the teacher is updated with an exponential moving average of the student encoder weights.
The model also contains a decoder trained using Flow Matching.

\subsection{Image modeling}
Self-supervised learning in image modeling includes approaches like DINO~\cite{caron_emerging_2021,oquab_dinov2_2024}, iBOT~\cite{zhou_ibot_2022}, SwAV~\cite{caron_unsupervised_2020}, BEiT~\cite{bao_beit_2022,peng_beit_2022,wang_image_2023}, and SimCLR~\cite{chen_simple_2020}. 
SimCLR~\cite{chen_simple_2020} uses two views of an image processed by a model which is trained to maximize agreement between the two outputs.
BEiT~\cite{bao_beit_2022} employs masked image modeling to pre-train a Transformer, later fine-tuned for classification and segmentation. 
It initially uses a Discrete Variational AutoEncoder (dVAE) for masked training labels, which is replaced in BEiT v2~\cite{peng_beit_2022} and BEiT-3~\cite{wang_image_2023} by the VQ-KD model. 
DiT~\cite{li_dit_2022} applies a similar strategy to document images.
In DINO~\cite{caron_emerging_2021,oquab_dinov2_2024}, a teacher model creates targets for training a student model.
The student is then trained using standard backpropagation, while the teacher is updated  with an exponential moving average of the student weights.
A similar approach to DINO is data2vec 2.0~\cite{baevski_efficient_2023}, where models are pre-trained in a very similar way on speech, images, and text.

\subsection{Text Recognition}
Self-supervised pre-training for text recognition primarily focuses on contrastive learning for scene text recognition~\cite{aberdam_sequence--sequence_2021,yang_reading_2022,guan_self-supervised_2023} and learning degradation-invariant models~\cite{souibgui_text-diae_2023}.
SeqCLR~\cite{aberdam_sequence--sequence_2021} processes two augmented views of a word image using noise contrastive estimation loss, while DiG~\cite{yang_reading_2022} employs InfoNCE loss and adds a reconstruction head to recover the masked original image. 
CCD~\cite{guan_self-supervised_2023} extends these methods with character-based segmentation, improving contrastive learning under geometric transformations.
In MaskOCR~\cite{lyu_maskocr_2023}, the encoder is pre-trained as a Context AutoEncoder and the decoder is trained predicting masked characters in synthetic text lines.
Text-DIAE~\cite{souibgui_text-diae_2023} trains an AutoEncoder to reconstruct degraded images, with its encoder later adapted for handwritten and scene text recognition.
Penarrubia et al.~\cite{penarrubia_spatial_2025} used rotation, flipping, jigsawing, and sorting to pretrain a CNN for text recognition model.
In SSM~\cite{gao_self-supervised_2024}, using augmentations two views of an image are combined into a single image and the model is pre-trained via reconstructing the original views.

\section{Masked Self-Supervised Pre-Training}
\label{sec:method}
\begin{figure}[t]
    \centering
    \begin{subfigure}{0.3\linewidth}
        \centering
        \includegraphics[width=\linewidth]{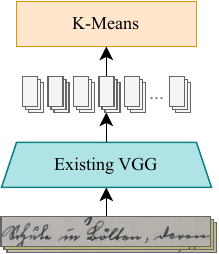}
        \caption{Fitting K-Means}
        \label{fig:method1}
    \end{subfigure}
    \hfill
    \begin{subfigure}{0.6\linewidth}
        \centering
        \includegraphics[width=\linewidth]{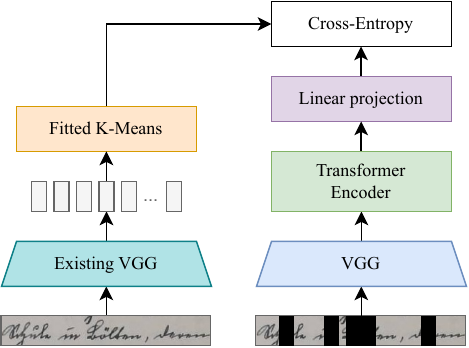}
        \caption{Pre-training}
        \label{fig:method2}
    \end{subfigure}
    \caption{Details of the masked self-supervised pre-training.
    We process text lines using existing VGG-like model to extract visual features and we fit a K-Means on them (Figure~\ref{fig:method1}).
    In the pre-training, we employ a VGG-like model and the fitted K-Means to generate discrete labels for a given text line. 
    Random patches of the text line are masked, and the encoder is trained to predict labels of the masked parts.
    (Figure~\ref{fig:method2}).}
    \label{fig:method}
\end{figure}

Our masked self-supervised pre-training is based on the masked label prediction where random patches in the input text line are masked and the model is trained to predict the class for these masked patches.
In particular, our method is based on the Feature Quantization~\cite{kiss_self-supervised_2024} method which utilizes existing feature extractor to encode patches of text lines into feature vectors that are used to fit a K-Means model.
Then, the K-Means model is used to obtain discrete labels for feature vectors obtained using the feature extractor.
The entire pre-training schema is visualized in Figure~\ref{fig:method}.

We propose two modifications to the pre-training phase.
Originally, the masking was done with a constant probability during the whole pre-training.
We propose to progressively increase the masking probability during training.
The reasoning behind this approach is that as a larger portion of the input image is masked, the task becomes more challenging for the model due to reduced available information.
This increased difficulty is expected to drive the model toward improved performance.
The second proposed modification changes the loss function which is originally calculated only on the masked patches.
We propose to compute the loss function also on the non-masked patches, to make better use of the information in the input data and to make the training more efficient, as the model learns on the entire text line, not only on a small subset.

\section{Experiments}
\label{sec:experiments}
We experimentally demonstrate the benefits of the masked self-supervised pre-training method when training encoder-decoder Transformers.
We perform our experiments on four various and differently sized datasets to show the robustness of the pre-training.
In the experiments, we follow the traditional self-supervised scheme, where the encoder is first pre-trained and then the whole model is fine-tuned on the downstream task using standard supervised training.
For the evaluation and comparison of the trained models, we use the character error rate (CER) metric.

In the first set of experiments, we evaluate the proposed pre-training strategies -- the progressive increase of masking probability and the calculation of the loss also on the non-masked patches.
In the second set, we focus on strategies during the fine-tuning phase.
Specifically, we compare two strategies where either (1) the entire model is optimized or (2) only the decoder is optimized first.
The motivation for the second strategy is that the pre-trained encoder should already generate meaningful outputs for given text lines and therefore it may be beneficial and efficient to train only the decoder.
The third set of experiments evaluates the performance of six differently sized pre-trained models with those trained from scratch.
In the last set, we compare the pre-trained models with models fine-tuned from existing OCR models, i.e. trained using transfer learning, which is a strong baseline in this kind of setup~\cite{kiss_self-supervised_2024}.

\subsection{Datasets}
In our experiments, we use two types of text line datasets.
The first type is unlabeled and it is used during the pre-training phase.
The second type is annotated with line-level transcriptions and it is used during the fine-tuning phase.

\paragraph{Unlabeled dataset}
The unlabeled dataset is composed of 30M text lines detected in 408k pages from 3k historical documents provided by archive in Opava\footnote{https://www.archives.cz/} and 20M text lines detected in 247k pages from obtained from our PERO-OCR web application\footnote{https://pero-ocr.fit.vutbr.cz/}.
In both cases, the documents contain mainly Czech and German handwritten pages from the late 19th century to the present.
The documents are mostly from chronicles, parish records, land registers, and similar.
The collection from the OCR application also contains some printed documents.
To detect text lines in these documents, we used ParseNet-based~\cite{kodym_page_2021} network trained on our internal layout dataset with annotated text lines.

\begin{figure}[t]
    \centering
    \begin{subfigure}{0.48\linewidth}
        \centering
        \includegraphics[width=\linewidth]{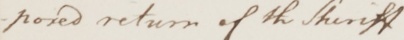}
        \caption{Bentham dataset}
        \label{fig:datasets_bentham}
        \vspace{0.3cm}
    \end{subfigure}
    \hfill
    \begin{subfigure}{0.48\linewidth}
        \centering
        \includegraphics[width=\linewidth]{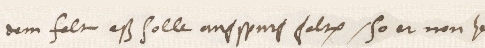}
        \caption{Bullinger dataset}
        \label{fig:datasets_dta}
        \vspace{0.3cm}
    \end{subfigure}
    \\ 
    \begin{subfigure}{0.48\linewidth}
        \centering
        \includegraphics[width=\linewidth]{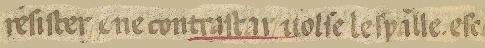}
        \caption{CATMuS Medieval}
        \label{fig:datasets_pero-printed}
        \vspace{0.3cm}
    \end{subfigure}
    \hfill
    \begin{subfigure}{0.48\linewidth}
        \centering
        \includegraphics[width=\linewidth]{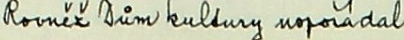}
        \caption{PERO-HWR}
        \label{fig:datasets_pero-hwr}
        \vspace{0.3cm}
    \end{subfigure}
    \\
    \begin{subfigure}{1.0\linewidth}
        \centering
        \includegraphics[width=0.48\linewidth]{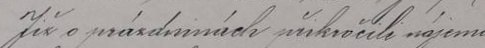}
        \hfill
        \includegraphics[width=0.48\linewidth]{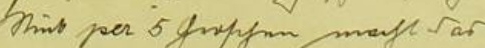}
        \caption{Unlabeled dataset}
        \label{fig:datasets_unlabeled}
    \end{subfigure}
    \caption{Examples of text lines from datasets.}
    \label{fig:datasets}
\end{figure}

\paragraph{Fine-tuning datasets} 
For the fine-tuning phase, we use four differently sized datasets: three standard, publicly available datasets and an internal handwritten dataset called \emph{PERO-HWR}.
The three standard datasets are: the ICFHR 2014 Bentham collection~\cite{sanchez_icfhr2014_2014}, The Bullinger Dataset~\cite{scius-bertrand_bullinger_2023}, and the CATMuS Medieval dataset~\cite{clerice_catmus_2024}.
The Bentham collection contains pages written in historical English by Jeremy Bentham and a few other authors.
The Bullinger dataset contains text lines from letter correspondences written in German and Latin.
The letters were either written by Heinrich Bullinger or addressed to him.
Unfortunately, we were not able to obtain the dataset as it is described in the original research paper because some parts with training text lines are not in the data source\footnote{https://github.com/pstroe/bullinger-htr}
The downloaded dataset also contains more text lines in the validation and the test sets than they report in the paper.
We pre-processed the obtained dataset using our text line detector resulting in text lines with more consistent baseline placement and text size.
The CATMuS Medieval dataset contains pages of historical documents spanning from the 8th to the 16th century written in 10 different languages.
We use the general split of the CATMuS Medieval v1.0 dataset.
The PERO-HWR dataset has similar properties to the unlabeled dataset -- it contains primarily Czech and German handwritten text lines from the late 19th century to the presetnt.
To a lesser extent, it also contains text lines from other existing datasets, such as ICDAR2017 READ dataset~\cite{sanchez_icdar2017_2017}, the Parzival dataset~\cite{fischer_lexicon-free_2012}, or the Saint-Gall dataset~\cite{fischer_transcription_2011}.
The sizes of all datasets are summarized in Table~\ref{tab:datasets} and their examples are shown in Figure~\ref{fig:datasets}.

\begin{table}[t]
    \centering
    \caption{%
        Sizes of the used datasets in the number of lines.
    }\label{tab:datasets}
    \begin{tabular}{
        @{\extracolsep{4pt}}@{\kern\tabcolsep}
        lrrr}
        \toprule    

        Dataset
        & \multicolumn{1}{c}{Training}
        & \multicolumn{1}{c}{Validation}
        & \multicolumn{1}{c}{Test}
        \\
        
        \midrule
        Unlabeled                      & 50\,573\,532 &    --   &  2\,000 \\ 
        \midrule
        Bentham                        &       9\,194 &  1\,415 &     860 \\ \addlinespace[0.1cm]
        Bullinger                      &      97\,442 & 16\,567 & 16\,567 \\ \addlinespace[0.1cm]
        CATMuS Medieval                &     146\,885 &  8\,317 &  8\,317 \\ \addlinespace[0.1cm]
        PERO-HWR   &  1\,149\,995 &    7\,383   &  -- \\ 
        
        \bottomrule
    \end{tabular}
\end{table}

\subsection{Preparation of the self-supervised dataset}
\begin{figure}[t]
    \centering
    \includegraphics[width=0.33\linewidth]{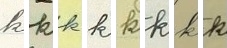}
    \hspace{1cm}
    \includegraphics[width=0.33\linewidth]{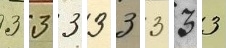}
    \vspace{0.1cm} \\
    \includegraphics[width=0.33\linewidth]{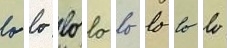}
    \hspace{1cm}
    \includegraphics[width=0.33\linewidth]{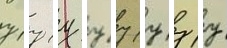}
    \vspace{0.1cm} \\
    \includegraphics[width=0.33\linewidth]{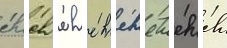}
    \hspace{1cm}
    \includegraphics[width=0.33\linewidth]{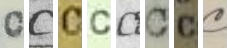}
    
    \caption{Visualization of identically encoded patches.
    Each group of patches in each row represent patches with the same trigrams (label triplets).}
    \label{fig:similarities}
\end{figure}

For the self-supervised dataset, we use the text lines from the unlabeled dataset.
The text lines are processed using an existing VGG-like model, producing visual features.
We obtained the model from a previously trained OCR on a previous version of the PERO-HWR dataset.
We fit a K-Means model with $4\,096$ classes on visual features of 15k randomly selected text lines, and we then produce discrete labels for the whole dataset.
To qualitatively evaluate the resulting labels, we visualized frequent trigrams in the dataset, as proposed by Kišš~et~al.~\cite{kiss_self-supervised_2024}.
Example visualizations are depicted in Figure~\ref{fig:similarities}.
Visualizations show that visually similar patches of text lines are encoded in the same trigrams (label triplets), which is essential from a pre-training perspective.

\subsection{Models}
\begin{figure}[t]
    \centering
    \includegraphics{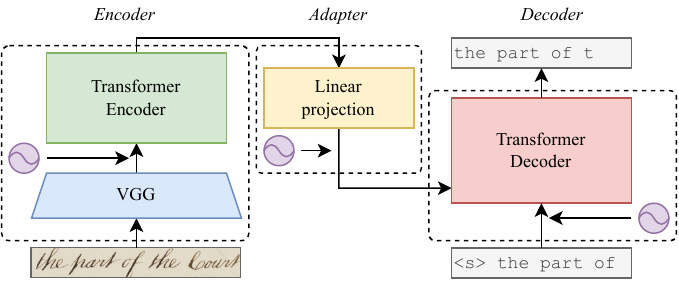}
    \caption{Schema of our models consisting of an encoder, a decoder, and an adapter that transforms the encoder outputs for the decoder.
    The purple cicles represent 1D sine-based positional encoding.}
    \label{fig:model}
\end{figure}

In the experiments, we use encoder-decoder Transformer architecture operating on arbitrarily long text lines with a normalized height of 48 pixels.
The visualization of the entire model is shown in Figure~\ref{fig:model}.
In the encoder, we use a VGG-like convolutional neural network, which transforms the input into a sequence of feature vectors with a horizontal subsampling factor of $8$.
The feature vectors are further processed with the standard Transformer encoder layers.
The decoder is a standard Transformer decoder consisting of an embedding layer translating the transcription characters into vectors, standard Transformer decoder layers, and an output projection producing categorical distribution over used alphabet.
The decoder is trained to generate the output sequence in an autoregressive manner.
In both, the encoder and the decoder, we add 1D positional information to the vectors before the Transformer layers using sine-based positional encoding.

Since we experiment with encoders and decoders of different dimensionality, we also have a small block, which we call an adapter (the middle part in Figure~\ref{fig:model}), that transforms the encoder output into a representation with correct dimensionality for the decoder.
The adapter is a linear layer with input and output channels equal to the dimensionality of the encoder and decoder, respectively.
After the linear layer, we also add 1D sine-based positional encoding.
In our preliminary experiments, this has proven to be a good practice when training a text recognizer with a pre-trained encoder, as the encoder does not necessarily retain positional information in its outputs.
Without this positional information, the decoder is prone to skipping or repeating sequences in its output.
On the other hand, it does not negatively affect performance of models without pre-trained encoder.

As mentioned before, we use encoders and decoders of different sizes.
In particular, we define two encoders, marked as E6 and E12, and three decoders, marked as D2, D6, and D10.
In all encoders and decoders, the MLP expansion ratio in the attention layers is $4$.
Details about the properties of the encoders and the decoders are summarized in Table~\ref{tab:encoders_decoders}.

\begin{table}[t]
    \centering
    \caption{%
        Definitions of the encoders and decoders.
    }\label{tab:encoders_decoders}
    \begin{tabular}{
        @{\extracolsep{4pt}}@{\kern\tabcolsep}
        lrrrr}
        \toprule    

        Name        & Layers & Heads &   Dim. & Params \\        
        \midrule
        Encoder E6  &      6 &     8 &    512 &   25 M \\ \addlinespace[0.1cm]
        Encoder E12 &     12 &    16 & 1\,024 &  158 M \\
        \midrule
        Decoder D2  &      2 &     4 &    320 &  3.6 M \\ \addlinespace[0.1cm]
        Decoder D6  &      6 &     8 &    512 &   26 M \\ \addlinespace[0.1cm]
        Decoder D10 &     10 &    12 &    768 &   95 M \\ 
        
        \bottomrule
    \end{tabular}
\end{table}

\subsection{Training details}
\paragraph{Pre-training phase}
In the pre-training phase, we optimize the encoders (the left part in Figure~\ref{fig:model}) with batch size 128 for 500k iterations.
The initial learning rate is set to $2\times 10^{-4}$ and it is halved after 100k and 300k iterations.
The optimized model consists of the convolutional encoder, Transformer encoder layers, and a linear projection layer, which transforms the output of the Transformer encoder layers into the classes produced by the K-Means model.
This projection layer is discarded from the model once the pre-training phase is finished.
Since the training targets are discrete labels, we train the encoders using standard cross-entropy loss.
For the pre-training, we use the code available in \texttt{pero-pretraining} github repository\footnote{https://github.com/DCGM/pero-pretraining}.

\paragraph{Fine-tuning phase}
In the fine-tuning phase, we train the models with batch size 64 using the standard cross-entropy loss.
We train the models for a different number of iterations depending on the dataset: 100k on the Bentham dataset, 500k on the CATMuS Medieval dataset and the Bullinger dataset, and 1M iterations on the PERO-HWR dataset.
The initial learning rate depends on the size of the model: we use $5\times10^{-5}$ for the encoder E6 and $1\times10^{-5}$ for the encoder E12.
When training on the Bentham dataset, the learning rate is halved after 30k iterations, on the Bullinger dataset and the CATMuS medieval dataset after 250k iterations, and on the PERO-HWR dataset after 500k iterations.

In our experiment, we also evaluate fine-tuning strategy in which we first train only the adapter and the decoder.
We train these two parts for the first 20 \% of iterations -- we refer to this as the decoder stage -- and then we train the entire model for the rest of the iterations.
Otherwise, the training setup remains the same.

In both the pre-training and fine-tuning phases, we train the models using bfloat16.
During the pre-training phase we do not use any augmentations as these would change the appearance of text lines and the self-supervised labels could be potentially wrong.
In the fine-tuning phase, we use augmentations consisting of color changing, adding noise, gamma adjusting, motion and defocus blurring, geometric transformations, and masking as proposed in AT-ST~\cite{kiss_at-st_2021}.

\paragraph{Transfer Learning}
In the transfer learning experiments, we train the models for 150k iterations with learning rate $2.5\times10^{-5}$.
Since we only perform these experiments on the Bullinger and the CATMuS datasets with encoder E6 and decoder D6, we only consider this single setup.
Otherwise, the training setup remains the same as in the fine-tuning phase.

\subsection{Pre-training experiments}
The first set of experiments is focused on the pre-training phase.
Kišš~et~al.~\cite{kiss_self-supervised_2024} randomly mask $20\,\%$ of the text lines.
In our experiments, we use the same strategy, and we also propose two modifications: progressive masking probability and training on non-masked patches.
In the first case, we increase the masking probability during training, in particular from the 100k-th iteration, we mask $33\,\%$ of the text line, and from the 300k-th iteration, we mask $50\,\%$ of the text line.
In the latter modification, we calculate the cross-entropy loss also on patches which were not masked and we add this to the loss calculated on the masked patches with weight $0.1$.

We evaluate the pre-training strategies by fine-tuning a model consisting of the smaller pre-trained encoder E6 and randomly initialized adapter and decoder D6 on the four text recognition datasets, measuring the character error rate of the fine-tuned models on the respective validation sets.
The results of the fine-tuned models are summarized in Table~\ref{tab:pretraining_results}.
The results show that the pre-training strategy with the progressive masking probability performs the best -- it consistently achieves the lowest CER on all datasets, except on the PERO-HWR dataset where the results are very close for each strategy.
Throughout the rest of the experiments, we use only encoders pre-trained using the progressive masking probability strategy.

\begin{table}[t]
    \centering
    \caption{%
        Results of models (E6+D6) with encoders pre-trained using the proposed strategies.
        We report the results in CER [\%] for validation sets of the respective fine-tuning datasets after the fine-tuning phase.
    }\label{tab:pretraining_results}
    \begin{tabular}{
        @{\extracolsep{4pt}}@{\kern\tabcolsep}
        cccccc}
        \toprule
        \multirow{2}{*}{\begin{tabular}[c]{@{}c@{}}\addlinespace[0.1cm] Progressive\\ masking prob.\end{tabular}} &
        \multirow{2}{*}{\begin{tabular}[c]{@{}c@{}}\addlinespace[0.1cm] Non-masked\\ patches loss\end{tabular}}
        & \multicolumn{4}{c}{Dataset} \\ \addlinespace[0.05cm] \cline{3-6} \addlinespace[0.15cm]
                   & & Bentham & Bullinger & CATMuS & PERO-HWR \\        
        \midrule
                  &           &   3.41  &     7.39  &  3.84  &  1.94  \\ \addlinespace[0.1cm] 
        \ding{51} &           &  \textbf{3.13}  &  \textbf{7.21}  &  \textbf{3.45}  &  1.94  \\ \addlinespace[0.1cm]
                  & \ding{51} &   3.57  &     7.29  &  3.49  &  \textbf{1.91}  \\ \addlinespace[0.1cm]
        \ding{51} & \ding{51} &   3.24  &     \textbf{7.21}  &  3.82  &  1.94  \\
        \bottomrule
    \end{tabular}
\end{table}

\subsection{Fine-tuning strategy experiments}
In this set of experiments, we evaluate two fine-tuning strategies.
The first strategy was already used in the previous set of experiments and it optimizes the entire model.
The second strategy first optimizes only the decoder and the adapter in the so-called decoder stage (first 20 \% of the training iterations) and later optimizes the entire model for the rest of the iterations.
As mentioned in the beginning of this section, the motivation behind this strategy is based on the idea that the pre-trained encoder should already be capable of producing meaningful outputs for given text lines, so it could be useful to train only the rest of the model, which is randomly initialized in the beginning of the training.
We used the same combination of encoder and decoder (E6 + D6) as in the previous set of experiments and we also report the character error rates on the respective validation sets.
The results of these experiments are in Table~\ref{tab:finetuning_results} and they show that training only the decoder in the decoder stage is not beneficial.
This also suggests that the encoder does not produce output directly suitable for the decoder that would allow simple downstream fine-tuning for text recognition.

\begin{table}[t]
    \centering
    \caption{%
        Results of models (E6+D6) with pre-trained encoders fine-tuned with the investigated fine-tuning strategies.
        We report the results in CER [\%] for validation sets of the respective fine-tuning datasets after the decoder stage of training (upper part) and for the best-performing checkpoint after the entire training (lower part).
    }\label{tab:finetuning_results}
    \begin{tabular}{
        @{\extracolsep{4pt}}@{\kern\tabcolsep}
        @{\hspace{1.5em}}lcccc}
        \toprule    

        \multirowcell{2}{Strategy} & \multicolumn{4}{c}{Dataset} \\ \cline{2-5} \addlinespace[0.15cm]
                    & Bentham & Bullinger & CATMuS & PERO-HWR \\        
        \midrule

        \multicolumn{5}{l}{\textit{After 20 \% iterations (decoder stage)}} \\ \addlinespace[0.1cm] 
        Decoder &  17.71  &  15.30  &  9.81  &  4.21  \\ \addlinespace[0.1cm] 
        Full    &   3.64  &   8.14  &  3.81  &  2.56  \\
        
        \midrule
        
        \multicolumn{5}{l}{\textit{Best performance after entire training}} \\ \addlinespace[0.1cm] 
        Decoder             &   3.70  &   7.88  &  4.16  &  \textbf{1.93}  \\ \addlinespace[0.1cm]
        Full                &  \textbf{3.13}  &  \textbf{7.21}  &  \textbf{3.45}  &  1.94  \\ \addlinespace[0.1cm]
        \bottomrule
    \end{tabular}
\end{table}

\subsection{Model size experiments}
\begin{figure}[t]
    \centering
    \begin{subfigure}{0.249\linewidth}
        \centering
        \includegraphics[width=\linewidth]{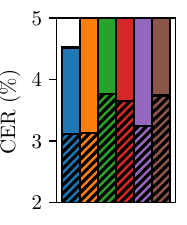}
        \caption{Bentham}
        \label{fig:results_sizes_bentham}
        \vspace{0.3cm}
    \end{subfigure}
    \hfill
    \begin{subfigure}{0.24\linewidth}
        \centering
        \includegraphics[width=\linewidth]{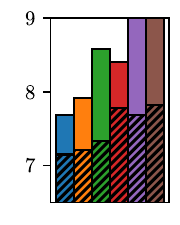}
        \caption{Bullinger}
        \label{fig:results_sizes_bullinger}
        \vspace{0.3cm}
    \end{subfigure}
    \hfill
    \begin{subfigure}{0.24\linewidth}
        \centering
        \includegraphics[width=\linewidth]{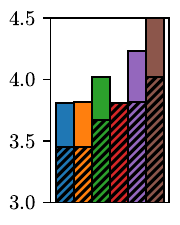}
        \caption{CATMuS}
        \label{fig:results_sizes_catmus}
        \vspace{0.3cm}
    \end{subfigure}
    \hfill
    \begin{subfigure}{0.24\linewidth}
        \centering
        \includegraphics[width=\linewidth]{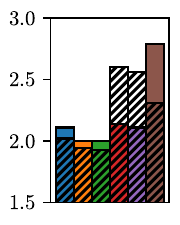}
        \caption{PERO-HWR}
        \label{fig:results_sizes_pero}
        \vspace{0.3cm}
    \end{subfigure}
    \includegraphics[height=1.2cm]{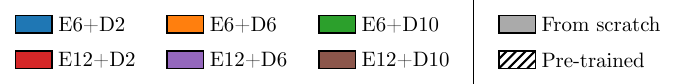}
    \caption{
    Results of pre-trained models and models trained from scratch on validations sets of the fine-tuning datasets in CER [\%].
    The different colors represent different models.
    Colored bars are results of the models trained from scratch and the hatched parts present results of pre-trained models.
    Note the different scales on each y-axis.}
    \label{fig:results_sizes}
\end{figure}

The third set of experiments is focused on the different sizes of encoders and decoders.
We use the E6 and E12 encoders pre-trained using the progressive masking probability strategy from the first set of experiments.
As decoders we use the D2, D6, and D10 decoders, and we optimize the entire encoder-decoder model as resulted from the second set of experiments.
A visual comparison of the performance of the models is shown in Figure~\ref{fig:results_sizes}.
It shows that the self-supervised pre-training helps to train better-performing models, especially on the smaller dataset.
It also turns out that training smaller models on these datasets is a better choice than training large models, even with pre-training.
On the other hand, the results for the PERO-HWR dataset show rather small improvements for the models with encoder E6.
There is even quite large increase of CER for the decoders D2 and D6 with the encoder E12.
It is necessary to note that we experienced instabilities during training of these models, which could probably be overcome by better hyper-parameter selection.
However, for the biggest model (encoder E12 + decoder D10) the pre-trained model again performed better than the one trained from scratch.
The particular CER values of the best-performing models on each dataset are in Table~\ref{tab:sizes_results}

\begin{table}[t]
    \centering
    \caption{%
        Results of the best-performing models from the experiments with model sizes.
        We report the results in CER [\%] for validation sets of the respective fine-tuning datasets.
        Under the name of each dataset, the configuration of the best-performing model is specified.
    }\label{tab:sizes_results}
    \begin{tabular}{
        @{\extracolsep{4pt}}@{\kern\tabcolsep}
        lcccc}
        \toprule    

        \multirowcell{3}{Training method} & \multicolumn{4}{c}{Dataset} \\ \cline{2-5} \addlinespace[0.15cm]
                    & Bentham & Bullinger & CATMuS  & PERO-HWR \\        
                    & (E6+D2) & (E6+D2)   & (E6+D2) & (E6+D10)    \\
        \midrule
        From scratch                    &   4.52  &   7.68  &  3.72  &  2.00  \\ \addlinespace[0.1cm]
        Self-supervised pre-training    &   \textbf{3.12}  &  \textbf{7.15}  &  \textbf{3.45}  &  \textbf{1.93}  \\        
        \bottomrule
    \end{tabular}
\end{table}

\subsection{Transfer learning experiments}
In this set of experiments, we used the E6+D6 model trained from scratch on the PERO-HWR dataset and we fine-tuned it on the Bullinger and CATMuS Medieval datasets to compare the self-supervised pre-training and transfer learning.
We did not run this experiment on the Bentham dataset because it is part of the PERO-HWR handwritten dataset, but unfortunately it does not follow the original splits, i.e. text lines from the original validation and test subsets are in the training subset of the PERO-HWR dataset.
Therefore, a fair comparison would not be possible.
The results are in Table~\ref{tab:transfer_learning} and they show that both transfer learning and self-supervised pre-training are beneficial when compared to training from scratch.
Moreover, the results of the self-supervisely pre-trained models achieve comparable or even slightly better results when compared to the transfer learning, which is usually considered a strong baseline~\cite{kiss_self-supervised_2024}.

\begin{table}[t]
    \centering
    \caption{%
        Results of the models (E6+D6) trained from scratch, using transfer learning, and using self-supervised pre-training.
        We report the results in CER [\%] for validation sets of the Bullinger and the CATMuS datasets.
    }\label{tab:transfer_learning}
    \begin{tabular}{
        @{\extracolsep{4pt}}@{\kern\tabcolsep}
        lcc}
        \toprule    

        \multirowcell{2}{Training method} & \multicolumn{2}{c}{Dataset} \\ \cline{2-3} \addlinespace[0.15cm]
                    & Bullinger & CATMuS \\
        \midrule
        From scratch                 & 7.91 & 3.90 \\ \addlinespace[0.1cm]
        Transfer learning            & \textbf{7.20} & 3.59 \\ \addlinespace[0.1cm]
        Self-supervised pre-training & 7.21 & \textbf{3.45} \\
        \bottomrule
    \end{tabular}
\end{table}

\subsection{Final comparison}
In Table~\ref{tab:final_results}, we present the final results on test sets of the Bentham, the Bullinger, and the CATMuS Medieval datasets.
In particular, we present the results of (1) the best-performing models (E6+D2) trained from scratch and using self-supervised pre-training, (2) models E6+D6 used through the experiments including transfer learning, and (3) state-of-the-art results for the Bentham dataset~\cite{parres_speed-up_2024} and the CATMuS Medieval dataset~\cite{clerice_catmus_2024} -- we do not report state-of-the-art results for the Bullinger dataset since we do not have the exact validation and test sets for fair comparison, as mentioned in the beginning of this section.
In the case of state-of-the-art results for the Bentham dataset, we report CER obtained by fine-tuning pre-trained TrOCR model on the Bentham, thus not really comparable with models trained from scratch, but comparable rather with the self-supervisedly pre-trained models.
The state-of-the-art results on the CATMuS Medieval dataset are obtained by training only on that dataset and thus these results are directly comparable to our models trained from scratch.
In each case, we selected the model with the lowest CER on the respective validation set and we report CER for the respective test set.

\begin{table}[t]
    \centering
    \caption{%
        Final results on test sets of the respective datasets.
        We report the results in CER [\%] for the best-performing models (upper part), models of configuration E6+D6 used through the experiments including transfer learning, which are thus mutually comparable in terms of size (middle part), and state-of-the-art results (lower part).
    }\label{tab:final_results}
    \begin{tabular}{
        @{\extracolsep{4pt}}@{\kern\tabcolsep}
        @{\hspace{1.5em}}lccc}
        \toprule    

        \multirowcell{2}{Training method} & \multicolumn{3}{c}{Dataset} \\ \cline{2-4} \addlinespace[0.15cm]
                    & Bentham & Bullinger & CATMuS  \\
        \midrule
        \multicolumn{4}{l}{\textit{Best models (E6+D2)}} \\
        From scratch                 & 3.86 & 7.62 & 3.72 \\ \addlinespace[0.1cm]
        Self-supervised pre-training & 2.51 & 7.08 & 3.49 \\
        \midrule
        \multicolumn{4}{l}{\textit{Comparable models (E6+D6)}} \\
        From scratch                 & 9.54 & 7.81 & 3.94 \\ \addlinespace[0.1cm]
        Transfer learning            &  --  & 7.11 & 3.63 \\ \addlinespace[0.1cm]
        Self-supervised pre-training & 2.47 & 7.12 & 3.49 \\
        \midrule
        State-of-the-art results     & 2.70 &  --  & 4.70 \\
        
        \bottomrule
    \end{tabular}
\end{table}

\section{Conclusions}
In this paper, we explored masked self-supervised pre-training for text recognition transformers, leveraging large-scale unlabeled data to improve recognition performance.
We proposed two modifications to the pre-training phase: progressively increasing the masking probability during training and incorporating non-masked patches into the loss computation.
Our experimental results demonstrate that the progressive masking modification leads to improved model performance after fine-tuning on the text recognition task.
Through extensive experiments on six differently sized models and four fine-tuning datasets, we showed that self-supervised pre-training on a large-scale unlabeled dataset provides a strong foundation for text recognition, particularly in scenarios where labeled data is limited.
Furthermore, our approach proved competitive with transfer learning, which is considered a strong baseline.

Future work may explore additional enhancements to the masked pre-training resulting in learning better representations.
Another option is investigating self-supervised pre-training with different models which could further improve the text recognition accuracy.
Additionaly, it could be beneficial either to pre-train the entire encoder-decoder model or to use pre-trained decoders.

\begin{credits}
\subsubsection{\ackname}
This work has been supported by the Ministry of Culture Czech Republic in NAKI III projects semANT - Semantic Document Exploration (DH23P03OVV060), Orbis Pictus – book revival for cultural and creative sectors (DH23P03OVV033), and Smart digiline - machine learning for digitization of printed heritage (DH23P03OVV066).

\subsubsection{\discintname}
The authors have no competing interests to declare that are
relevant to the content of this article.
\end{credits}

\bibliographystyle{splncs04}
\bibliography{references}

\end{document}